\documentclass{ieeeaccess}
\usepackage{cite}
\usepackage{amsmath,amssymb,amsfonts}
\usepackage{algorithm}
\usepackage{algorithmic}
\usepackage{graphicx}
\usepackage{textcomp}
\usepackage{booktabs}
\usepackage{array}

\usepackage{microtype}
\usepackage{graphicx}
\usepackage{booktabs} 
\usepackage{subfig}

\usepackage{xspace}
\usepackage{amsmath}
\usepackage{amssymb}
\usepackage{amsfonts}
\usepackage{bm}
\usepackage{bbm}
\usepackage{float}
\usepackage{stfloats}
\usepackage{wrapfig}
\usepackage[pagebackref=true,breaklinks=true,colorlinks,bookmarks=false]{hyperref}
\usepackage{multirow}
\usepackage{lipsum}
\usepackage{mathtools}
\usepackage{cuted}
\usepackage{stfloats}

\definecolor{Green}{RGB}{102,252,102}

\newcommand{\ie}{\textit{i.e. }}
\newcommand{\TP}[1]{\textcolor{black}{#1}}

\newcommand{\PreserveBackslash}[1]{\let\temp=\\#1\let\\=\temp}
\newcolumntype{C}[1]{>{\centering\arraybackslash}p{#1}}

\def\BibTeX{{\rm B\kern-.05em{\sc i\kern-.025em b}\kern-.08em
    T\kern-.1667em\lower.7ex\hbox{E}\kern-.125emX}}

\begin{document}
\history{23 May 2024}
\doi{10.1109/ACCESS.2024.3404474}

\title{Uncertainty-Aware Rank-One MIMO Q Network Framework for Accelerated Offline Reinforcement Learning}

\author{\uppercase{Thanh Nguyen }\authorrefmark{1},  \uppercase{Tung Luu}\authorrefmark{1}, \uppercase{Tri Ton}\authorrefmark{1}, \uppercase{Sungwoong Kim}\authorrefmark{2} and \uppercase{Chang D. Yoo \authorrefmark{1}}}

\address[1]{School of Electrical Engineering, Korea Advanced Institute of Science and Technology, Daejeon 34141, Republic of Korea}
\address[2]{Department of Artificial Intelligence, Korea University, Seoul 02841, Republic of Korea}

\markboth
{Author \headeretal: Preparation of Papers for IEEE TRANSACTIONS and JOURNALS}
{Author \headeretal: Preparation of Papers for IEEE TRANSACTIONS and JOURNALS}

\corresp{Corresponding author: Chang D. Yoo (e-mail: cd\_yoo@kaist.ac.kr).}

\begin{abstract}
\TP{Offline reinforcement learning (RL) has garnered significant interest due to its safe and easily scalable paradigm, which essentially requires training policies from pre-collected datasets without the need for additional environment interaction. However, training under this paradigm presents its own challenge: the extrapolation error stemming from out-of-distribution (OOD) data}. Existing methodologies have endeavored to address this issue through means like penalizing OOD Q-values or imposing similarity constraints on the learned policy and the behavior policy. Nonetheless, these approaches are often beset by limitations \TP{such as being overly conservative in utilizing OOD data}, imprecise OOD data characterization, and significant computational overhead. \TP{To address these challenges, this paper introduces an Uncertainty-Aware Rank-One Multi-Input Multi-Output (MIMO) Q Network framework. The framework aims to enhance Offline Reinforcement Learning by fully leveraging the potential of OOD data while still ensuring efficiency in the learning process. 
Specifically, the framework quantifies data uncertainty and harnesses it in the training losses, aiming to} train a policy that maximizes the lower confidence bound of the corresponding Q-function. Furthermore, a Rank-One MIMO architecture is introduced to model the uncertainty-aware Q-function, \TP{offering the same ability for uncertainty quantification as an ensemble of networks but with a cost nearly equivalent to that of a single network}. Consequently, this framework strikes a harmonious balance between precision, speed, and  memory efficiency, culminating in improved overall performance. Extensive experimentation on the D4RL benchmark demonstrates that the framework attains state-of-the-art performance while remaining computationally efficient. By incorporating the concept of uncertainty quantification, our framework offers a promising avenue to alleviate extrapolation errors and enhance the efficiency of offline RL.

\end{abstract}


\titlepgskip=-15pt

\maketitle

\section{Introduction}

Offline reinforcement learning (RL), also known as batch RL, addresses the challenge of training a policy solely from a fixed dataset without additional interaction with the environment. This approach offers benefits in terms of data efficiency, scalability, and safety, particularly in real-world applications such as navigation and healthcare \cite{kumar2019stabilizing,farahmand2011action,kumar2020conservative}, leading to a drastically increasing attention in recent years \cite{fujimoto2018addressing,fujimoto2019off,kostrikov2021offlinefis}. However, learning from \TP{pre-collected} datasets \TP{ presents a well-known challenge: the extrapolation error when performing policy evaluation on out-of-distribution (OOD) data.} Specifically, estimating the \TP{Q-values} of the learned policy using the offline dataset is prone to bias due to the distributional shift caused by differences in visitation distribution between the learned policy and the behavior policy used for collecting the dataset. \TP{This bias, once combined with deep network fitting in deep RL algorithms, leads to a significant extrapolation error characterized by the Q-function's significant overestimation of OOD data, as evidenced in \cite{fujimoto2019off}. Addressing this error in offline RL proves challenging due to the absence of interaction for collecting additional data, a process common in conventional RL methods, to rectify it.}

\TP{Confronted with the inherent limitations of the problem, offline RL algorithms overcome the challenge by proposing different losses or training procedures capable of mitigating extrapolation errors. Early methods directly limit the distributional shift by constraining the learned policy to be similar to the behavior policy \cite{fujimoto2018addressing,fujimoto2019off,fujimoto2021minimalist,wu2019behavior}. This approach has a notable drawback: the learned policy is heavily influenced by the behavior policies, leading to suboptimal results in datasets collected by non-optimal behavior ones. Later methods adopt a different approach by making conservative estimates of future rewards, aiming to learn a value function that serves as a strict lower bound to the true value function \cite{kumar2019stabilizing, kumar2020conservative, kostrikov2021offlinefis}. Typically, this involves penalizing Q-functions for OOD actions. While these methods offer more freedom for learning better policies, the penalizing term often lacks precise characterization of OOD data (e.g. equally
penalizes the OOD actions), resulting in overly conservative value functions \cite{buckman2020importance}. Recent methods}, considering uncertainty about the value function \cite{buckman2020importance,jin2021pessimism,xie2021bellman}, have been proposed to address the excessively pessimistic nature of the aforementioned approaches. \TP{These methods commonly employ separate Q-functions as a Q-ensemble, enabling uncertainty-aware penalization. By penalizing conflicting actions and favoring decisions that are consistent across the models, they generate pessimistic Q-values to train the learned policy \cite{agarwal2020optimistic,fujimoto2018addressing,bai2022pessimistic,yang2022rorl}, and have proven} to achieve state-of-the-art performances. However, the use of separate Q-functions incurs high computational and memory complexity. Moreover, these methods introduce additional hyperparameters and require a large number of ensemble members to be effective, posing difficulties for large and complex datasets.

\TP{Taking into account the shortcomings of prior approaches as a whole, which include a conservative use of OOD data, imprecise characterization of such data, and substantial computational overhead,  we propose the end-to-end Uncertainty-Aware Rank-One MIMO Q Network framework for improving offline RL.} Our framework fully leverages uncertainty quantification to effectively utilize OOD data, thereby enhancing the learning process while ensuring fast training and memory efficiency. 

\TP{Our contribution can be listed in detail as follows:}
\begin{itemize}
    \item Firstly, \TP{we introduce a novel architecture}, \ie Rank-One MIMO Q network, \TP{for approximating} the naive Q ensemble. \TP{The Rank-One MIMO Q network combines a common shared network with mini rank-one network adapters, allowing these adapters to fuse the common shared network to assemble ensemble members. This network can handle multiple inputs and generate multiple outputs simultaneously. As a result, the proposed network offers the same capability for uncertainty quantification as an ensemble of networks but with a cost nearly equivalent to that of a single network. This result is achieved by recognizing} that members of an ensemble can collectively acquire and retain certain common knowledge about the environment, thereby eliminating the necessity for individual learning and storage. \TP{In our design, within each layer, the MIMO Q network stores shared knowledge in the shared network weight matrix. Individual members can then augment this shared knowledge with their unique insights} using their independent weights, modeled by two vectors. \TP{This design minimizes} the parameter overhead of our MIMO Q compared to the naive ensemble, thanks to the utilization of a shared body. \TP{Furthermore, the MIMO Q is facilitated with matrix vectorization to enable uncertainty prediction in a single forward pass, optimizing prediction speed efficiency.}

    \item Secondly, \TP{we propose pessimistic training losses as a means to leverage uncertainty quantification for effectively utilizing OOD data. These losses are} based on maximizing the lower confidence bounds (LCB) of Q values, which have been proven to be a precise characterization of OOD data. \TP{Specifically, our proposed losses are designed to train} the MIMO Q network and the policy network based on the conservatively estimated Q value suggested by the min-valued MIMO Q head. \TP{This design} enables efficient estimates of the LCB of Q-values with only one hyper-parameter, the number of Q heads, for controlling pessimism. \TP{Furthermore, the backward pass only needs to propagate through the min-valued Q member instead of all the ensemble members, as conventionally done in the naive ensemble method. This further enhances the speed of the learning process.} 
    
    \item Thirdly, \TP{we} enhance training stability and mitigate overestimation without the need for an OOD sampling scheme \TP{by incorporating} two components: \TP{(1)} maximizing entropy for OOD actions while maximizing likelihood for in-distribution actions. \TP{It plays a crucial role in preventing excessive exploitation of OOD actions, thereby reducing the risk of diverging Q-values, while simultaneously encouraging the exploitation of trustworthy in-distribution actions.} These additions are particularly effective for low-coverage datasets, such as expert datasets;  \TP{(2) employing} a lazy policy improvement trick. This not only saves computational costs but also enhances the stability of policy evaluation.
    \item  \TP{Finally, we conduct extensive experimental results and rigorous ablation studies. The experimental result demonstrates} that the framework achieves state-of-the-art performance on the D4RL benchmark \cite{fu2020d4rl} while being computationally friendly compared to strong baselines. The ablation studies \TP{are carefully chosen to provide} a better understanding of the framework. 
\end{itemize}

\TP{Broadly}, this work highlights the importance of developing efficient, stable ensembling techniques specifically designed for offline RL. \TP{Additionally, it underscores} the potential of offline RL as a testbed for validating uncertainty estimation techniques and raises intriguing research questions for future exploration.

The paper is organized as follows: Section 1 introduces the topic, followed by Section 2, which delves into related work. Section 3 provides background information on offline RL, and Section 4 thoroughly outlines our methodology, encompassing the Rank-One MIMO Q network and the Uncertainty-Aware Rank-One MIMO Q Network framework. Section 5 presents the experiment setup. Section 6 shows the experimental results, and Section 7 conducts an ablation study. Finally, Section 8 draws the paper to a conclusion and explores potential avenues for future research.

\begin{figure*}[htp]
\centering
\includegraphics[width=0.8\textwidth]{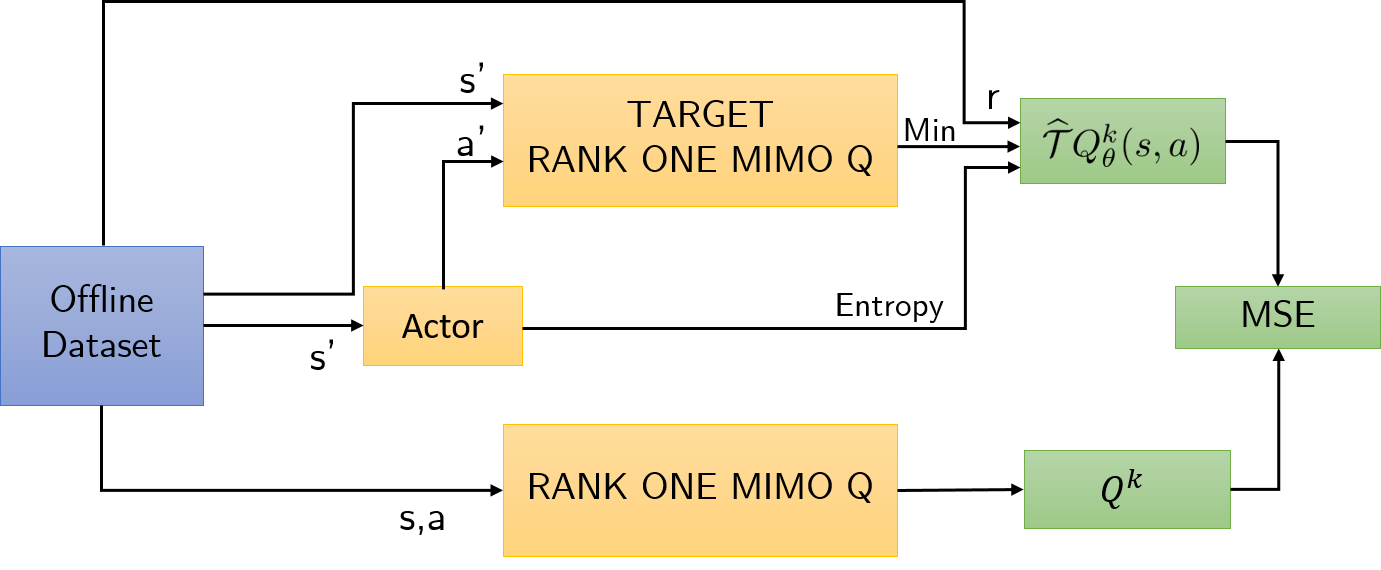}
\caption{Illustration of the proposed workflow: Our framework adopts a conventional actor-critic setup, where the online Q network and target Q network are alternated by the proposed Rank-One MIMO Q, while the policy is modeled by a stochastic Gaussian actor. As the framework prohibits direct interaction with the environment, transitions are sampled from a pre-collected offline dataset. These sampled transitions are then utilized to calculate the target, employing the minimum-valued head of the target network and policy entropy. Finally, the calculated target is backed up to the online network using mean square error during training.}
\label{fig:architecure}
\end{figure*}

\section{Related Work}

\TP{\textbf{Offline Reinforcement Learning:} This paper primarily delves into model-free offline Reinforcement Learning (RL). Early methods pinpoint the core issue in offline RL as extrapolation error \cite{fujimoto2019off} and suggest using policy constraints to ensure that the learned policy remains close to the behavior policy}. These constraints include adding behavior cloning (BC) loss \cite{torabi2018behavioral} in policy training \cite{fujimoto2021minimalist}, using the divergence between the behavior policy and the learned policy\cite{fujimoto2018addressing,fujimoto2019off,kumar2019stabilizing}, applying advantage-weighted constraints to balance BC and advantages\cite{peng2019advantage}, penalizing the prediction-error of a variational auto-encoder\cite{rezaeifar2022offline}, and learning latent actions from the offline data\cite{zhou2021plas}. \TP{While policy-constraint methods excel in performance on datasets derived from expert behavior policies, they struggle to discover optimal policies when confronted with datasets featuring suboptimal policies. This limitation arises from the stringent constraints imposed on the learned policies}\cite{lee2021sunrise,nair2020awac}. \TP{Furthermore, these methods necessitate precise estimation of the behavior policy, a task typically challenging in complex settings characterized by multiple sources of behavior or high-dimensional environments. Subsequent methods circumvent these limitations by instead learning a pessimistic Q-function,  which serves as a lower bound estimate of the true value function.  This is performed by penalizing their Q-value \cite{kumar2020conservative, kumar2019stabilizing,cheng2022adversarially} or using V-learning \cite{ma2021offline,kostrikov2021offline}.  While offering increased flexibility for improved policy learning, the penalizing terms in these methods often lack a precise characterization of OOD data. For instance, CQL \cite{kumar2020conservative} uniformly penalizes the Q-values of all OOD samples, a practice proven to result in conservative value functions, as demonstrated in \cite{buckman2020importance}. Our method overcomes the limitations of previous approaches by adopting uncertainty quantification to selectively penalize OOD data. While our approach shares similarities with CQL \cite{kumar2020conservative} in promoting conservatism in Q-learning, it distinguishes itself by explicitly measuring the uncertainty of OOD actions rather than applying a uniform penalty.}

\TP{It's worth noting that uncertainty quantification has been used in online RL \cite{bai2021principled, sekar2020planning, azizzadenesheli2018efficient, mavrin2019distributional, nikolov2018information, yang2022rorl}}. However, offline RL poses \TP{their own} challenge \TP{for} uncertainty quantification \TP{compared to online RL, primarily} due to the limited coverage of offline data and the distribution shift of learned policies. 
\TP{Several existing model-based and model-free approaches have been proposed to overcome this challenge}. In model-based offline RL, representative works include MOPO \cite{yu2020mopo} and MOReL \cite{kidambi2020morel}, which utilize an ensemble dynamics model for uncertainty quantification, while BOPAH \cite{lee2020batch} combines uncertainty penalization with behavior-policy constraints. \TP{However, existing model-based methods suffer from additional computation costs and perform sub-optimally in complex environments \cite{chua2018deep,janner2019trust}. In contrast, our work conducts model-free learning, which is less affected by these problems. In addition, model-free learning is also considered more favorable due to its simplicity and potential for high performance. To be noticed, while there are similar approaches proposed in model-free offline RL, our work stands apart from them. To be specific,}  UWAC \cite{wu2021uncertainty} uses dropout-based uncertainty for model-free offline RL but relies on policy constraints for value function learning \cite{gal2016dropout}. In comparison, our method with the proposed MIMO Q network demonstrates greater robustness to distributional shifts compared to the ensemble and dropout-based approaches. EDAC \cite{an2021uncertainty} also employs ensemble Q networks, but it diversifies gradients to penalize OOD actions, and PBRL \cite{bai2022pessimistic} penalizes OOD actions through direct OOD sampling and associated uncertainty quantification, whereas our method doesn't require further OOD sampling.

\TP{\textbf{Efficient Uncertainty Quantification in Supervised Learning}: Our work draws inspiration from recent advancements in efficient uncertainty quantification in Supervised Learning to identify appropriate approaches. It is well-known that} ensembles have demonstrated their effectiveness in estimating uncertainty in RL, just as they have in supervised deep learning \cite{smit2021pebl,an2021uncertainty,bai2022pessimistic}. Despite the availability of other technical methods for uncertainty estimation \cite{li2007robust,pawlowski2017implicit}, ensembles consistently outperform them, albeit at a higher computational cost. Therefore, in supervised learning, much of the current research on ensembles is aimed at improving computational efficiency, with proposals for reducing compute or memory footprint during training and inference on large ensembles \cite{havasi2020training,wen2020batchensemble,liu2021deep}. Several architectures have been proposed to enable shared knowledge among ensemble members. The \TP{multi-head approach \cite{lee2015m,osband2016deep,tran2020hydra} designs ensembles that share a big “trunk” network and have separate “head” networks for each ensemble member. While multi-head approaches offer reduced computational costs compared to typical ensembles by sharing many layers, they often lack ensemble diversity \cite{havasi2020training}.  On the other hand, MIMO \cite{havasi2020training} overcomes this problem by modifying both the input and output layers to be a multi-input multi-output network, allowing each ensemble member to take different paths throughout the full network. However, MIMO networks are reported to struggle with fitting more than 2 subnetworks \cite{rame2021mixmo}, necessitating special care for the shared body layers \cite{rame2021mixmo,sun2022towards}. Our work shares similarities with the MIMO approach, an improved version of the multi-head approach. However, we explicitly model shared knowledge and each ensemble member individually, a strategy that has been proven to be effective for fitting multiple sub-networks \cite{wen2020batchensemble}.
Beyond the domain of supervised learning, within the realm of offline reinforcement learning,} there is limited empirical evidence showcasing the effectiveness of these approaches. Inspired by efficient methods in supervised learning, our work aims to leverage these techniques to benefit offline reinforcement learning.

\section{Background}

Offline reinforcement learning (RL) is a research field that \TP{distinguishes itself from} online RL \cite{mnih2013playing,nguyen2021sample,nguyen2021robust,luu2022utilizing,luu2022visual,nguyen2024robust} by enabling training without the need for active interaction with the environment. This involves utilizing a fixed dataset, collected by an unknown behavior policy, to learn a policy that maximizes the cumulative reward of a target environment. By leveraging previously collected data, offline RL offers a more practical and efficient solution for RL training in complex domains where online interaction is not feasible or too expensive.

This paper considers the fully-observed Markov Decision Process (MDP) as commonly utilized in offline RL research. Mathematically, the MDP is defined by a tuple $\mathcal{M} = (\mathcal{S},\mathcal{A},\mathbb{P},\rho_0,R,\gamma, H)$. Therein, $\mathcal{S}$ is a set of state $s$, $\mathcal{A}$ is a set of actions $a$, $\mathbb{P}$ is the transition probability of the dynamics in the form $\mathbb{P}(s_{t+1}|s_t,a_t)$, $\rho_0$ is the initial state distribution, $R$ is reward function, $\gamma \in (0,1]$ is a scalar discount factor, and $H$ is the horizon. 

Within a MDP, offline RL try to learn a policy $\pi(a_t|s_t)$  which is the probability of taking action $a_t$ conditioned on the current state $s_t$. A trajectory distribution, which is a sequence of $H+1$ states and $H$ actions, can be further derived as $\tau = (s_0,a_0,...,s_H)$ where $H$ can be infinite. The probability density function for a given trajectory $\tau$ under policy $\pi$ is as below:\begin{equation}
p_{\pi}(\tau)=\rho_{0}\left(s_{0}\right) \prod_{t=0}^{H-1} \pi\left(a_{t} \mid s_{t}\right) \mathbb{P}\left(s_{t+1} \mid s_{t}, a_{t}\right).    
\end{equation}
In offline RL, a dataset comprising multiple pre-collected transitions is given, conveniently denoted as $D = \{(s, a, s^\prime, r)\}$. Here, $s^\prime$ represents the next state resulting from taking action $a$ at the current state $s$ and receiving a return $r$. This dataset is gathered by an unknown behavior policy $\pi_\beta$. The primary objective of offline RL is to learn an optimal $\pi^*$ policy that maximizes the expected cumulative return of the learned policy $\pi$. The objective can be formulated as follows:
\begin{equation}
\pi^{*}=\underset{\pi}{\operatorname{argmax}} \quad \mathbb{E}_{\tau \sim p_{\pi}}\left[\Sigma_{t=0}^{H-1} \gamma^{t} R\left(s_{t}, a_{t}\right)\right].
\end{equation}
 The optimal policy's corresponding Q-function satisfies the Bellman operator as shown below:
\begin{equation}
\mathcal{T} Q_\theta(s, a):=R(s, a)+\gamma \mathbb{E}_{s^{\prime} \sim \mathbb{P}(\cdot \mid s, a)}\left[\max _{a^{\prime}} Q_{\theta^{-}}\left(s^{\prime}, a^{\prime}\right)\right],    
\end{equation}
where $\theta$ represents the parameters of the Q network, and $\theta^{-}$ represents the parameters of the target-network, which is a copy of the Q network with momentum update for training stabilization \cite{mnih2015human}.

Offline RL poses challenges due to the distribution shift when training policies from a pre-collected dataset. Value function evaluated on the greedy action $a'$ in the Bellman operator $\mathcal{T} Q(s,a) = R(s,a) + \gamma \mathbb{E}_{s^{\prime}}\left[\max _{a^{\prime}} Q(s^{\prime}, a^{\prime})\right]$ tends to have extrapolation errors, as the combination of state-action pairs $(s', a')$ may have rarely occurred in the dataset $D$. To overcome this issue, early model-free offline RL methods incorporate conservatism by constraining learned policies to be similar to the behavior policy or penalizing values of OOD actions. However, these methods often limit the generalization of value functions beyond the offline data and lack precise characterization of OOD data. Uncertainty quantification is a promising way to enhance performance. Online RL typically uses upper-confidence bound (UCB) to encourage exploration, while offline RL focuses on fixed training data and uses lower-confidence bound (LCB) to estimate Q-values and avoid risky actions. Leveraging the LCB is supported by strong theoretical evidence in offline RL. Recent theoretical analyses, such as those discussed in \cite{jin2021pessimism, xie2021bellman}, prove the importance of uncertainty quantification in achieving provable efficiency in RL. Pessimistic Value Iteration \cite{jin2021pessimism} introduces an $\epsilon$-uncertainty quantifier, which serves as a penalty and enables provable efficient pessimism in offline RL. In the context of linear MDPs, a LCB-penalty \cite{abbasi2011improved,jin2020provably} is proposed and is a known $\epsilon$-uncertainty quantifier. In the context of function approximation, it is shown that the bootstrapped uncertainty provides an estimation of the LCB-penalty and enables efficient pessimism in many complex offline RL tasks \cite{bai2022pessimistic}. This is achieved by utilizing the Q learning form below, where the $K$ bootstrapped Q-functions in critic are used to quantify the epistemic uncertainty.
\begin{equation}
\begin{aligned}
&\widehat{\mathcal{T}} Q_\theta^k(s, a)= R(s, a)\\
&+\gamma \widehat{\mathbb{E}}_{s^{\prime} \sim D, a^{\prime} \sim \pi(\cdot \mid s)}\left[\bar{Q}_{\theta^{-}}\left(s^{\prime}, a^{\prime}\right)-\beta \mathcal{U}_{\theta^{-}}\left(s^{\prime}, a^{\prime}\right)\right]    
\end{aligned}
\label{eq:PBRL_Qobjective}
\end{equation}
\begin{equation}
\begin{aligned}
\mathcal{U}_{\theta^{-}}(s, a)&=\operatorname{Std}\left(Q_{\theta^{-}}^k(s, a)\right) \\
&= \sqrt{\frac{1}{K} \sum_{k=1}^K\left(Q_{\theta^{-}}^k(s, a)-\bar{Q_{\theta^{-}}}(s, a)\right)^2}.    
\end{aligned}
\end{equation}
In this formulation, $Q_\theta^k$ represents the $k^{th}$ Q-function in the ensemble of bootstrapped Q-functions, while $\bar{Q}$ is the mean among the target-networks. The empirical Bellman operator $\hat{\mathcal{T}}$ aims to estimate the expected maximum Q-value $\mathbb{E}\left[R(s, a)+\gamma\max_{a'} Q_{\theta^{-}}\left(s^{\prime}, a^{\prime}\right) \mid s, a\right]$ based on the offline dataset, as $a'$ is sampled from the learned policy that is designed to maximize the Q-function. The epistemic uncertainty, $\mathcal{U}$, is quantified by the deviation among the bootstrapped Q-functions, and is used as a penalization in estimating the Q-functions. From a Bayesian perspective, the ensemble approach allows for estimation of the posterior distribution of the Q-functions, yielding similar values in areas with rich data and diverse values in areas with scarce data. A potential drawback of this approach is its high computational cost, which can be attributed to the use of a naive ensemble and its non-efficient objective forms \cite{ghasemipour2022so, bai2022pessimistic,an2021uncertainty}.

\section{Methodology}

\textbf{Rank-One MIMO Q network (MIMO Q):} Inspired by recent work in efficient ensemble \cite{havasi2020training,wen2020batchensemble}, we introduce the MIMO Q for fast and memory-efficient Q ensembling. The philosophy behind our method is that members of the ensemble can share certain common knowledge about the environment, which need not be learned and stored individually. Instead, this knowledge can be acquired collectively and stored in a shared weight matrix, which individual members can then use to learn their own unique knowledge with their own individual weights. As a result, we can model the ensemble as the product of a shared matrix and a rank-one matrix personalized for each member, enabling us to combine both collective and individual learning in a powerful and efficient way.

\begin{figure}[htp]
\centering
\includegraphics[width=1.0\linewidth]{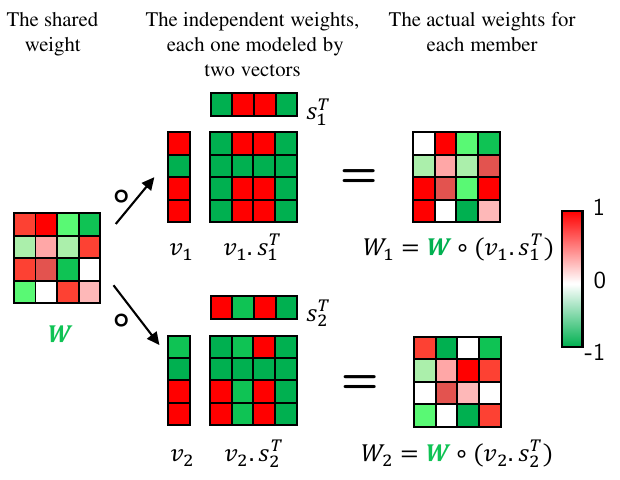}
\caption{Demonstration of how to generate ensemble weights for two ensemble members from one rank-one layer. It's important to note that the training weight stored in each rank-one layer consists of only the shared weight and two vectors for each member. The actual weights for each member are calculated on demand following equation \ref{rank_one_equation}. }
\label{fig:rank_one_process}
\end{figure}

\TP{Following this philosophy, we design the architecture of MIMO Q network accordingly. The MIMO Q network will function as an ensemble of $K$ members, as usual, meaning it will receive $K$ inputs and produce corresponding $K$ outputs. However, the MIMO Q network is constructed by stacking multiple special layers, referred to as rank-one layers, akin to building a multilayer perceptron from dense layers, rather than creating $K$ independent networks for each member in the ensemble. The pivotal aspect unfolds within the rank-one layer, meticulously crafted to model both a shared weight and $K$ individual weights. Specifically, the learnable weights for a rank-one layer consist of a shared weight that is common across all ensemble members, as well as $K$ independent weights that correspond individually to each of the $K$ ensemble members. Mathematically, the shared weight is essentially a conventional dense layer, denoted by $W \in \mathbb{R}^{m \times n}$, where $m$ represents the input dimension and $n$ represents the output dimension. On the other hand, one independent weight consists of a pair of vectors, $v \in \mathbb{R}^m$ and $s \in \mathbb{R}^n$, with dimensions $m$ and $n$ respectively. The layer comprises $K$ ensemble independent weights, implying that there are $K$ pairs of trainable vectors $v_k$ and $s_k$, for $k$ ranging from $1$ to $K$. During the forward pass, the rank-one layer constructs the actual ensemble weights $W_k$ by fusing the shared weight and the individual weights. This fusion is formulated using the following procedure:}
\begin{equation}
W_k=W \circ (v_k s_k^{\top}),
\label{rank_one_equation}
\end{equation}
where $\circ $ is element-wise multiplication and $(v_k s_k^{\top})$ is the dot product to recover the rank-one matrix from two vectors.  \TP{Figure \ref{fig:rank_one_process} visualizes the process of creating actual weight for each member of the ensemble having two members.}

To make the ensemble weight generation process parallelizable on a device, vectorization can be used. This allows the forward pass to be computed with respect to multiple ensemble members at once. The computation is accomplished through manipulation of matrix computations for a mini-batch \cite{wen2018flipout}. Specifically, let $x$ represent the \TP{feature map inputing to a} neural network layer. The \TP{output feature map, $y$,} are then given by:
\begin{equation}
\begin{aligned}
y_{bk} & =\Phi\left(W_k^{\top} x_b\right) \\
& =\Phi\left(\left(W \circ v_k s_k^{\top}\right)^{\top} x_b\right) \\
& =\Phi\left(\left(W^{\top}\left(x_b \circ v_k\right)\right) \circ s_k\right),
\end{aligned}    
\end{equation}
where $\Phi$ denotes the activation function and the subscript $b$ represents the index in the mini-batch. The output, \TP{$y_{bk}$}, represents \TP{next layer’s feature map input} from the $k^{th}$ ensemble member. To enable vectorization of these computations, matrices $V$ and $S$ are defined such that their rows consist of the vectors $v_k$ and $s_k$ for all examples in the mini-batch. With this, the equation above can be expressed in a vectorized form as follows:
\begin{equation}
Y=\Phi(((X \circ V) W) \circ S),
\label{eq:linearRankoneLayer}
\end{equation}
where $X$ is the mini-batch input. By computing equation \ref{eq:linearRankoneLayer}, we can obtain the next layer’s \TP{feature map input} for each ensemble member in a mini-batch-friendly way. This allows us to take full advantage of parallel accelerators to implement the ensemble efficiently. To match the input and the ensemble weight, we can divide the input mini-batch into $K$ sub-batches and each sub-batch receives an ensemble weight $W_k, k=\{1, \ldots, K\}$ or we can repeat the input mini-batch $K$ times.

For evaluation, when the test batch size is $B$ and there are $K$ ensemble members, we optimize efficiency by repeating the input mini-batch $K$ times, resulting in an effective batch size of $B \times K$. This allows all ensemble members to compute the output for the same $B$ input data points in a single forward pass.

\TP{Diversity in the weights of ensemble members is a crucial factor for an ensemble to achieve high performance and reduce the number of members required for effective uncertainty quantification}. Encouraging diversity during training can be achieved through the use of a loss function (e.g., diversity loss) \cite{an2021uncertainty}, but this approach incurs additional computational costs. Instead, we adopt an alternative method by initializing individual weights as random sign vectors \cite{wen2020batchensemble}, which has been found to yield satisfactory results without incurring any extra computational overhead.

\textbf{Uncertainty-Aware Rank-One MIMO Q Network framework:} Considering PBRL as the baseline, the proposed framework based on bootstrapped uncertainty quantification builds on the actor-critic scheme, consisting of policy evaluation and policy improvement phases. The Uncertainty-Aware Rank-One MIMO Q Network framework additionally modifies both the policy evaluation loss and the policy improvement loss to include the benefits of the framework.

\TP{In our policy evaluation, we integrate two key components: an effective term designed to approximate the lower confidence bound (LCB) of the Q-value predictions, and a value bonus derived from OOD action entropy. The proposed Bellman equation is as follows:}

\begin{equation}
\begin{aligned}
&\widehat{\mathcal{T}} Q_\theta^k(s, a)=R(s, a)\\
&+\gamma \widehat{\mathbb{E}}_{s^{\prime} \sim D, a^\prime \sim \pi(\cdot \mid s^{\prime})}\left[\min _{k=1, \ldots, K} Q_{\theta^{-}}^k\left(s^{\prime}, a^{\prime}\right) -\alpha \log \pi_\phi\left(a^{\prime} \mid s^{\prime}\right)\right]. 
\end{aligned}
\label{equa:minapproximation}
\end{equation}

Let's delve into the details of each component. Addressing extrapolation error using the uncertainty quantification approach is done by maximizing the LCB of the Q-values. Instead of using equation \ref{eq:PBRL_Qobjective} to compute the expected target, our framework chooses the worst-case Q-value, which can be interpreted as an approximation of the LCB of the Q-value predictions. Mathematically, suppose $Q(s, a)$ follows a Gaussian distribution with mean $\mu$ and standard deviation $\sigma$. Let \TP{ $\left\{Q^k(s, a)\right\}_{k=1}^K$ be realizations of $Q(s, a)$. }The expected minimum of the realizations can be approximated using the work of Royston \cite{royston1982expected}, which also employed in \cite{an2021uncertainty,fujimoto2021minimalist,haarnoja2018soft,fujimoto2018addressing} as:
\begin{equation}
 \mathbb{E}\left[\min _{k=1, \ldots, K} Q^k(s, a)\right] \approx \mu(s, a)-\mathcal{F}^{-1}\left(\frac{K-\frac{\pi}{8}}{K-\frac{\pi}{4}+1}\right) \sigma(s, a).   
\end{equation}
In the equation, $\mathcal{F}$ denotes the cumulative distribution function of the standard Gaussian distribution. This relation reveals that using the minimum value approximates the penalty on the ensemble mean of the Q-values minus the standard deviation scaled by a coefficient that depends on the number of ensembles, denoted by $K$. This approximation enables the estimation of the LCB in a computationally efficient manner. Furthermore, in the backward pass, employing this approximation results in the loss being backpropagated solely through the minimum-value Q network rather than through all ensemble members. This property is particularly advantageous in the context of our framework, as its computational costs remain insensitive to $K$.

Regarding the value bonus, the Q-function receives an additional boost from the entropy of actions generated by the learned policy in the next stage, which has a high probability of being OOD data. This approach encourages the Q-function to avoid over-reliance on any specific high-value OOD data, thereby preventing the exploitation of OOD data, which easily leads to over-estimation.

In the policy improvement, we use the combination of (1) minimum MIMO Q value, (2) the entropy of policy-generated action, and (3) the likelihood of action in the dataset to derive the corresponding policy by solving the following maximization problem:
\begin{equation}
\begin{aligned}
\pi_{\phi}&=\max _{\phi} \widehat{\mathbb{E}}_{s,a \sim \mathcal{D}, a' \sim \pi(\cdot \mid s)}[\min _{k=1, \ldots, K} Q^k(s, a')\\
&-\alpha \log \pi_\phi\left(a' \mid s\right) + \beta \log \pi_\phi\left(a \mid s\right)],       
\end{aligned}
\end{equation}
where $\phi$ represents the policy parameters. 

The technique of maximizing entropy is utilized to encourage a diverse set of actions in a given state. This is achieved by treating the policy as a probability distribution over actions given a state and maximizing its entropy, as done in SAC \cite{haarnoja2018soft}. By doing so, the policy becomes less deterministic, allowing the agent to explore more effectively and reducing the risk of overfitting the training data. Furthermore, maximizing entropy has a regularizing effect on the policy evaluation process. Encouraging the policy to take a diverse set of actions also reduces the risk of exploiting maximum actions (highly likely to be adversarial examples), which can lead to diverging Q-values in the long run. From another perspective, using a high-entropy Gaussian policy has the effect of smoothing out the Q target, which improves the robustness of the Q-function to outliers. Furthermore, in the context of offline RL, we must balance the use of in-distribution and out-of-distribution data, which respectively be sampled from the dataset and sampled from the learned policy. While OOD data can be useful for seeking a better optimistic result, in-distribution data is more trustworthy and should be given some priority. Therefore, the log-likelihood term in the loss function is designed to incentivize the policy to favor actions that are in distribution, ensuring a balanced and effective learning strategy. The maximization of the log-likelihood term is particularly useful in low-coverage environments, such as expert datasets. In high-coverage environments, this term may be less important and can be eliminated without any impact on performance. As a whole, our objective for policy improvement allows us to avoid penalizing OOD data using some additional high computational cost losses, as is done in some other approaches such as PBRL, CQL \cite{bai2022pessimistic,kumar2020conservative}. 

It is worth noting that updating the policy network too frequently not only incurs higher computational costs but can also introduce instability during policy evaluation. Therefore, in our approach, we update the policy network after a certain number of policy evaluation updates. This ensures a balance between computational efficiency and stability in the learning process.

The overall architecture of our framework is illustrated in figure \ref{fig:architecure}.

\begin{table*}[t]
\tabcolsep=0.11cm
\centering
\begin{tabular}{@{}rl|ccccccccc|c@{}}
\toprule
\multicolumn{2}{c|}{Task Name}               & \begin{tabular}[c]{@{}c@{}}BCQ \\ \cite{fujimoto2019off} \end{tabular}& \begin{tabular}[c]{@{}c@{}}IQL \\ \cite{kostrikov2021offline} \end{tabular} &\begin{tabular}[c]{@{}c@{}}BEAR \\ \cite{kumar2019stabilizing} \end{tabular}       & \begin{tabular}[c]{@{}c@{}}UWAV \\ \cite{wu2021uncertainty} \end{tabular}       & \begin{tabular}[c]{@{}c@{}}CQL \\ \cite{kumar2020conservative} \end{tabular}         &\begin{tabular}[c]{@{}c@{}} MOPO \\ \cite{yu2020mopo} \end{tabular}      & \begin{tabular}[c]{@{}c@{}}TD3-BC \\ \cite{fujimoto2021minimalist} \end{tabular}    & \begin{tabular}[c]{@{}c@{}}EDAC-10 \\ \cite{an2021uncertainty} \end{tabular}   & \begin{tabular}[c]{@{}c@{}}PBRL \\ \cite{bai2022pessimistic} \end{tabular}      & OURS        \\ \midrule
\multirow{3}{*}{\rotatebox{90}{Random}}        & HalfCheetah & 2.3 ±0.0 & 14.4 ±2.3 & 2.3 ±0.0    & 2.3 ±0.0   & 17.5 ±1.5   & 35.9 ±2.9  & 11.0 ±1.1  & 13.4 ± 1.1 & 11.0 ±5.8  & \textbf{31.5±2.3}   \\
                               & Hopper      & 10.5 ±0.2 & 11.1 ±0.1 & 3.9 ±2.3    & 2.7 ±0.3   & 7.9 ±0.4    & 16.7 ±12.2 & 8.5 ±0.6   & 16.9±10.1  & 26.8 ±9.3  & \textbf{31.3±0.6}   \\
                               & Walker2d    & 4.2 ±1.5 & 5.8 ±0.3 & 12.8 ±10.2  & 2.0 ±0.4   & 5.1 ±1.3    & 4.2 ±5.7   & 1.6 ±1.7   & 6.7±8.8    & 8.1 ±4.4   & \textbf{21.3±0.2}   \\ \midrule
\multirow{3}{*}{\rotatebox{90}{Medium}}        & HalfCheetah & 40.1 ±0.5 & 41.7 ±6.2 & 43.0 ±0.2   & 42.2 ±0.4  & 47.0 ±0.5   & \textbf{73.1 ±2.4}  & 48.3 ±0.3  & 64.1±1.1   & 57.9 ±1.5  & 68.6±2.6   \\
                               & Hopper      & 52.5 ±22.8 & 64.1±0.3 &51.8 ±4.0   & 50.9 ±4.4  & 53.0 ±28.5  & 38.3 ±34.9 & 59.3 ±4.2  & \textbf{103.6±0.2}  & 75.3 ±31.2 &    93.2±7.4        \\
                               & Walker2d    & 47.7±7.8 & 78.3±0.6 &-0.2 ±0.1   & 75.4 ±3.0  & 73.3 ±17.7  & 41.2 ±30.8 & 83.7 ±2.1  & 87.6±11.0  & 89.6 ±0.7  &   \textbf{92.2±0.3}         \\ \midrule
\multirow{3}{*}{\rotatebox{90}{\begin{tabular}[c]{@{}c@{}}Medium\\ Replay\end{tabular}}} & HalfCheetah & 39.0±1.9 & 42.5±0.3 & 36.3 ±3.1   & 35.9 ±3.7  & 45.5 ±0.7   & 69.2 ±1.1  & 44.6 ±0.5  & 60.1±0.3   & 45.1 ±8.0  & \textbf{59.9±2.4}   \\
                               & Hopper      &11.1±7.7 & 81.9±17.9 & 52.2 ±19.3  & 25.3 ±1.7  & 88.7 ±12.9  & 32.7 ±9.4  & 60.9 ±18.8 & 102.8±0.3  & 100.6 ±1.0 & \textbf{102.9±0.7} \\
                               & Walker2d    &15.2±4.6 & 70.9±3.4 & 7.0 ±7.8    & 23.6 ±6.9  & 81.8 ±2.7   & 73.7 ±9.4  & 81.8 ±5.5  & 94.0±1.2   & 77.7 ±14.5 & \textbf{100.6±3.4} \\ \midrule
\multirow{3}{*}{\rotatebox{90}{\begin{tabular}[c]{@{}c@{}}Medium\\ Expert\end{tabular}}} & HalfCheetah & 60.1±11.1 & 75.4±3.5 &46.0 ±4.7   & 42.7 ±0.3  & 75.6 ±25.7  & 70.3 ±21.9 & 90.7 ±4.3  & \textbf{107.2±1.0}  & 92.3 ±1.1  & 100.9 ±3.3 \\
                               & Hopper      &110.5±2.8 & 91.5±0.2 & 50.6 ±25.3  & 44.9 ±8.1  & 105.6 ±12.9 & 60.6 ±32.5 & 98.0 ±9.4  & 58.1±22.3  & 110.8 ±0.8 & \textbf{111.6±0.7}  \\
                               & Walker2d    & 43.6±14.0 & 109.6±1.4 &22.1 ±44.9  & 96.5 ±9.1  & 107.9 ±1.6  & 77.4 ±27.9 & 110.1 ±0.5 & \textbf{115.4±0.5}  & 110.1 ±0.3 & 112.9±1.2 \\ \midrule
\multirow{3}{*}{\rotatebox{90}{Expert}}        & HalfCheetah &90.4±7.5 & 94.8±2.5 & 92.7 ±0.6   & 92.9 ±0.6  & 96.3 ±1.3   & 81.3 ±21.8 & 96.7 ±1.1  & 104.0±0.8  & 92.4 ±1.7  & \textbf{105.4±2.3}  \\
                               & Hopper      &103.6±5.0 & 106.2±8.9 & 54.6 ±21.0  & 110.5 ±0.5 & 96.5 ±28.0  & 62.5 ±29.0 & 107.8 ±7   & 77.0±43.9  & \textbf{110.5 ±0.4} & 107.8±0.9           \\
                               & Walker2d    &110.4±0.2 & 109.0±0.3 & 106.6 ±6.8  & 108.4 ±0.4 & 108.5 ±0.5  & 62.4 ±3.2  & 110.2 ±0.3 & 57.8±55.7  & 108.3 ±0.3 & \textbf{112.7±0.3} \\ \midrule
\multicolumn{2}{c|}{Average}                 &49.4 & 66.5 & 38.78  & 50.41  & 67.35   & 53.3  & 67.55  & 71.2  & 74.37  &   \textbf{83.6  }    \\ \bottomrule
\end{tabular}
\caption{The average normalized scores and standard deviations of all algorithms across four seeds in the D4RL benchmark \cite{fu2020d4rl}. The highest-performing scores are highlighted for each dataset. The state-of-the-art scores referenced in this study are from \cite{bai2022pessimistic,yang2022rorl}.}
\label{tab:benchmark}
\end{table*}

\section{Experimental Setup}

Our method is evaluated on the D4RL benchmark \cite{fu2020d4rl}, which consists of various continuous-control tasks and datasets. Specifically, we evaluate our method on three environments (HalfCheetah, Hopper, and Walker2d) in the Gym domain, using five types of datasets for each environment: random-v2, medium-v2, medium-replay-v2, medium-expert-v2, and expert-v2. The medium-replay dataset contains experiences collected during the training up to a medium-level policy, while the random/medium/expert dataset is generated by a single random/medium/expert policy. The medium-expert dataset is a combination of the medium and expert datasets.

In all experiments, we train our algorithms for 3000 epochs, which corresponds to 1000 training steps per epoch and a total of 3 million steps. This training setup and common hyper-parameters follow the guidelines of PBRL and EDAC \cite{bai2022pessimistic,an2021uncertainty}. The reported results are normalized using d4rl scores, which provide a measure of performance relative to expert and random scores. The normalization formula is as follows:
\begin{equation}
score_{normalized} = 100* \frac{score - score_{random} }{ score_{expert} - score_{random}}.  
\end{equation}
 For evaluation, each algorithm is assessed based on 10 trajectories, with 1000 steps per trajectory, and the returns are averaged over 4 random seeds.

For specific hyper-parameters, we employ an adaptive learning rate $\alpha$ based on (SAC) algorithm \cite{haarnoja2018soft}. The value of $\alpha$ is determined dynamically during training. The number of members in the ensemble, denoted as $K$, is selected through a random search within the range of 2 to 20. As for $\beta$, in the case of expert datasets, we perform a random search within the range of $[0, 1e3]$, while for other datasets, the random search is conducted within the range of $[0,1]$. We update the policy network every 2 policy evaluation updates to strike a balance between computational efficiency and stability.

\section{Experimental Result}

\textbf{Result on benchmark datasets:} We compare our method with several state-of-the-art algorithms: \TP{(1) BCQ, \cite{fujimoto2019off} aiming to mitigate extrapolation error by constraining the action space of the trained policy to be similar to the behavior policy, learned from the dataset, (2) IQL \cite{kostrikov2021offline}, adopting an implicit Q-learning approach where the Q-function is learned based on a learned V-function without directly querying a Q-function with OOD actions.}
(3) BEAR \cite{kumar2019stabilizing}, which enforces policy constraints using the Maximum Mean Discrepancy (MMD) distance, (4) UWAC \cite{wu2021uncertainty}, an improvement on BEAR that incorporates dropout uncertainty-weighted updates, (5) CQL \cite{kumar2020conservative}, which learns conservative value functions by minimizing Q-values of OOD actions, (6) MOPO \cite{yu2020mopo}, which quantifies uncertainty through ensemble dynamics in a model-based setting, (7) TD3-BC \cite{fujimoto2021minimalist}, which incorporates adaptive behavior cloning constraints to regularize the policy during training, (8) EDAC \cite{an2021uncertainty}, which utilizes a diversified Q-ensemble to enforce conservatism, (9) PBRL \cite{bai2022pessimistic}, which applies uncertainty penalization and OOD sampling. It is worth noting that EDAC and PBRL are related to our method since all these methods employ a Q-ensemble for conservatism.

Table \ref{tab:benchmark} reports the performance of the average normalized score with standard deviation. Based on the results, \TP{our framework achieves state-of-the-art performance, significantly surpasses other techniques. Especial, on average, our approach nearly doubles the scores achieved by methods such as BCQ and BEAR. Moreover, it outpaces the closest competitor (PBRL) by a substantial margin of +9.23. }

\TP{A detailed examination of individual tasks reveals that our framework demonstrates superior performance compared to, or is at least on par with, the previous strongest approach.} Notably, our framework exhibits significant improvements over \TP{the second-best result} with a large margin when handling messy data, such as random or medium replay.

The comparison of our method with several state-of-the-art algorithms in offline RL empirically proves several key advantages that contribute to our method's superior performance.

Firstly, let's discuss the contrast with model-based offline RL methods such as MOPO. While model-based approaches likely encounter difficulties in accurately estimating dynamics from data, leading to sub-optimality, our method operates independently of explicit dynamics estimation as it is model-free. This removes concerns associated with accurately estimating multi-modal data distributions. Additionally, model-free learning is recognized not only for its simplicity but also for its potential to achieve high performance.

Secondly, our method outperforms techniques that constrain the learned policy to be similar to the behavior policy, such as BCQ, BEAR, and TD3-BC. These methods impose stringent constraints on policy learning, tying the learned policy closely to a potentially suboptimal behavior policy, which might stem directly from suboptimal data or poor estimation of the behavior policy. In contrast, our method operates independently of any specific behavior policy estimate, allowing for more flexible and potentially more optimal policy learning.

Thirdly, in comparison to approaches utilizing pessimistic Q-functions like IQL and CQL, our method offers increased flexibility and precision in handling out-of-distribution (OOD) data. While pessimistic Q-functions often lack precise characterization of OOD data, our method explicitly measures the uncertainty of OOD actions and selectively penalizes them, rather than applying a uniform penalty. This targeted approach enhances the robustness and adaptability of our method, particularly in scenarios with diverse and messy data distributions.

Lastly, when compared to methods employing uncertainty quantification techniques like UWAC, PBRL, and ADAC, our approach excels in handling messy data, such as random or medium replay. 
This superiority can be attributed to our method's more accurate quantifying uncertainty, enabled by a MIMO Q network architecture.  This architecture fosters effective sharing of common knowledge among network members while allowing enough flexibility for each member to discover their own insights without the burden of individual learning from scratch.

Upon the observations, we believe that for an offline RL method to be deemed the best, it should demonstrate several key qualities: avoiding errors from dynamics model estimation (e.g: using model-free RL), flexible policy optimization independent of behavior policy constraints, superior uncertainty quantification capabilities, and selective penalization of out-of-distribution (OOD) data. Our method exemplifies these qualities, enabling it to achieve state-of-the-art performance on benchmark tasks.

\begin{figure}[htp]
\centering
\includegraphics[width=0.45\textwidth]{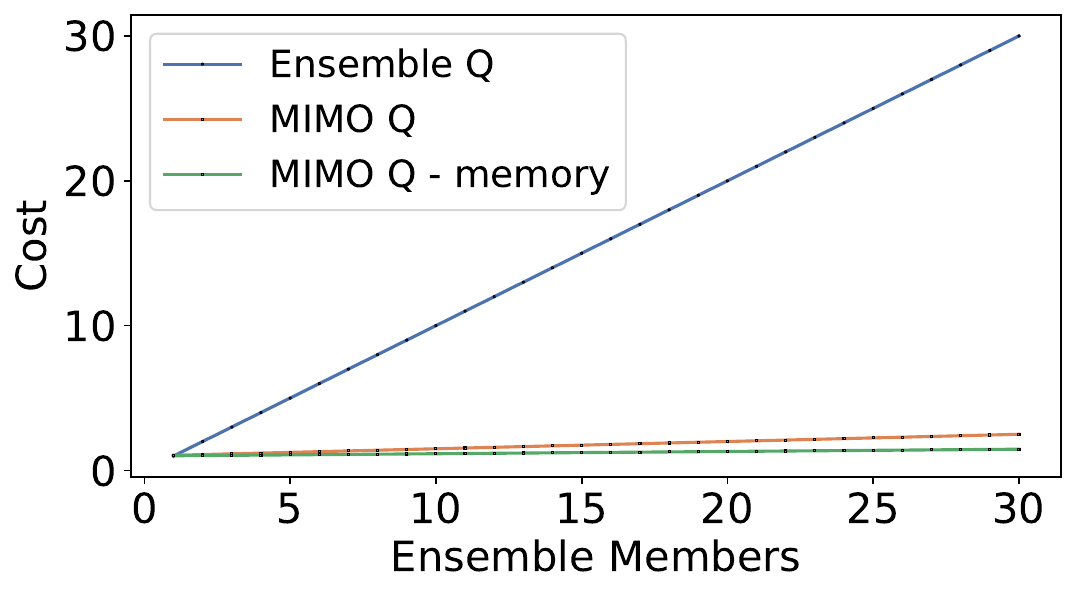}
\caption{The forward time cost and memory of MIMO Q w.r.t the ensemble size. The result is relative to the single model cost.}
\label{fig:ensemble_cost}
\end{figure}

\textbf{Time and space complexity:} The naive ensemble relies on employing separate models for each member, resulting in linear increases in forward cost and memory cost as the ensemble size grows. In contrast, our approach utilizes the shared weight across the ensemble. \TP{The only extra weights introduced are individual weights, which incur an extremely small overhead thanks to the utilization of rank-one vectors}. As a result, the cost remains nearly unchanged, \TP{equivalent to that of a single network,} even as the ensemble size increases as shown in Figure \ref{fig:ensemble_cost}. 

\begin{table}[htp]
\centering
\begin{tabular}{@{}l|cc@{}}
\toprule
     & \begin{tabular}[c]{@{}c@{}}Runtime\\ (s/epoch)\end{tabular} & \begin{tabular}[c]{@{}c@{}}GPU memory\\ (GB)\end{tabular} \\ \midrule
CQL  & 32.4                                                        & 1.4                                                       \\
PBRL & 102.96                                                      & 1.8                                                       \\
OUR  & 17.8                                                        & 0.97                                                      \\ \bottomrule
\end{tabular}
\caption{Computational costs on hopper-medium using Tesla V100}
\label{tab:runtime}
\end{table}

\TP{Mathematically, in our approach, the only additional memory required is for storing the sets of vectors {$v_1$, ..., $v_k$} and {$s_1$, ..., $s_k$}, which have low memory overhead compared to weight matrices. Assuming the Q network consists of $L$ fully connected layers with dimensions $m \times n$, the ensemble weight size is reduced from $LmnK$ in the naive ensemble to $Lmn + K(m + n)$. This reduction in memory usage is a significant advantage of our approach.}

To better understand the empirical efficiency of our framework, we present a runtime analysis in Table \ref{tab:runtime}. The results demonstrate that our method not only achieves the best runtime performance but also maintains exceptional memory efficiency. Remarkably, it operates 1.82 times faster than CQL and 5.87 times faster than PBRL. In terms of memory usage, our framework is the most efficient, requiring only 0.97GB, compared to CQL's 1.4GB and PBRL's 1.8GB. 

The reasons behind these efficiency gains are clear. Our method runs slightly faster than CQL, mainly because CQL's computational demands stem primarily from its reliance on out-of-distribution (OOD) action sampling and the logsumexp approximation. On the other hand, PBRL faces slow processing times due to its separate modeling of Q networks, the necessity of OOD action sampling, and the requirement for calculations from prior functions to enhance diversity. In stark contrast, our method operates without the need for OOD sampling. Additionally, it efficiently enhances diversity through a random sign vector initialization approach. Furthermore, we employ a clever lazy policy improvement technique, which not only reduces computational costs but also enhances the stability of policy evaluation.

In summary, our method not only outperforms other state-of-the-art approaches but also exhibits significantly greater efficiency in terms of both runtime and memory usage. These improvements represent substantial advancements in the field of offline RL.

\section{Ablation study}

\begin{figure}[htbp]
\centering
\includegraphics[width=0.5\textwidth]{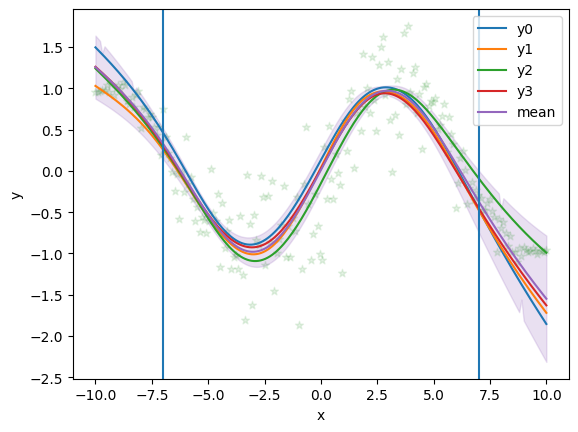}
\caption{An illustration of the behavior of the Rank-One Q Network on the synthetic regression dataset. $y_i$ denotes the prediction of each head. }
\label{fig:regressionResult}
\end{figure}

\textbf{Uncertainty quantification ability:} To gain a better understanding of the effectiveness of uncertainty quantification, we present a simple prediction task as an illustration. In this task, we train the Rank-One MIMO Q network using training data on the $\mathbb{R}^1$ plane, specifically within the range of $[-7.0, 7.0]$. The input variable $x$ is generated from a Gaussian-distributed sinusoid function. For test data, we extend the data range to [-10, 10], following the same underlying function. We visualize the testing data points along with the corresponding uncertainty quantification in figure \ref{fig:regressionResult}. As depicted in the figure, the uncertainty quantification gradually increases as we move from the in-distribution data points to the OOD data points. This visual representation demonstrates the Rank-one MIMO Q's ability to perform regression with reliable uncertainty quantification, even on OOD data. 

\textbf{The effect of $K$:} We conducted experiments by varying the parameter $K$ and recorded the corresponding outcomes in table \ref{tab:ablationK}. The experimental results confirm that adjusting the parameter $K$ effectively controls pessimistic behavior, aligning with the theoretical expectations. Lower values of $K$ tend to yield more optimistic results, while higher values of $K$ lead to greater pessimism. The optimal outcome lies in finding an approximate value of $K$ that strikes a balance between optimism and pessimism, resulting in the best overall performance.

\begin{table}[htpb]
    \centering
    \begin{tabular}{@{}l|ccccc@{}}
    \toprule
     $K $    & 2    & 5    & 10     & 15   & 20    \\ \midrule
    Avg Return  & 0.19 & 92.9 & 112.8  & 23   & 0.4   \\ \midrule
    Avg $Q(s,\pi(s))$ & 3e11 & 410  & 373.46 & -5e6 & -2e12 \\ \bottomrule
    \end{tabular}
    \caption{The effect of K on final performance and Q estimation on walker2d-medium-expert }
    \label{tab:ablationK}
\end{table}

\textbf{Component-wise Analysis:}  We provide experimental results of the framework when different components are omitted. The baseline model we used is the basic framework, which does not incorporate the entropy bonus or the in-distribution likelihood maximization. Table \ref{tab:component} presents the ablation results, indicating the impact of these additional components. It is evident that our framework achieves the best performance when leveraging both the entropy and in-distribution likelihood maximization.

Interestingly, both the entropy and likelihood terms introduce a certain level of pessimism, as evidenced by the corresponding average Q values of 267.1 and 258.9, respectively. On the other hand, not using either term yields a higher Q value. Nevertheless, the performance remains satisfactory even without the inclusion of these components. It is worth noting that in the case of expert datasets such as walker2d-expert, the inclusion of in-distribution likelihood maximization plays a crucial role. Without it, achieving good results becomes highly unstable and challenging. The in-distribution likelihood maximization component provides stability and enhances the performance of the model when working with expert datasets.

\begin{table}[htpb]
\centering
\begin{tabular}{@{}cccc@{}}
\toprule
\multicolumn{1}{l}{Entropy} & \multicolumn{1}{l}{Likelihood} & \multicolumn{1}{l}{Avg Return} & \multicolumn{1}{l}{Avg $Q(s,\pi(s))$} \\ \midrule
\checkmark                        & \checkmark                         & \textbf{112.9}                               & 373.4                        \\
                                 \checkmark &                          & 107.3                           & 267.1                       \\
                        & \checkmark                                    & 111.0                           & 258.9                       \\
                                  &                                    & 109.6                           & 453.2                       \\ \bottomrule
\end{tabular}
\caption{Component-wise analysis on the walker-2d medium-expert dataset. }
\label{tab:component}
\end{table}

\section{Conclusion}

In conclusion, this paper introduces a novel framework for offline reinforcement learning that utilizes uncertainty quantification to effectively leverage reliable OOD data. The proposed framework for precisely learning a policy through maximizing the lower confidence bound of the Q-function, along with the Rank-One Multi-Input Multi-Output architecture, strikes a balance between computational cost and precision, providing good overall performance. Extensive experiments on the D4RL benchmark show that our proposed framework achieves state-of-the-art performance while being computationally friendly.
Our findings contribute to the advancement of offline RL research and pave the way for future research in leveraging uncertainty quantification in an efficient way for addressing challenges in offline RL.

\section{Acknowledgement}

This work was supported by Institute for Information \& communications Technology Planning \& Evaluation (IITP) grant funded by the Korea government (MSIT) (No. 2021-0-01381, Development of Causal AI through Video Understanding and Reinforcement Learning, and its applications to real environments) and partly supported by Institute of Information \& communications Technology Planning \& Evaluation (IITP) grant funded by the Korea government (MSIT) (No. 2022-0-00951, Development of Uncertainty-Aware Agents Learning by Asking Questions).

\bibliographystyle{plain}
\bibliography{refs.bib}

\begin{IEEEbiography}[{\includegraphics[width=1in,height=1.25in,clip,keepaspectratio]{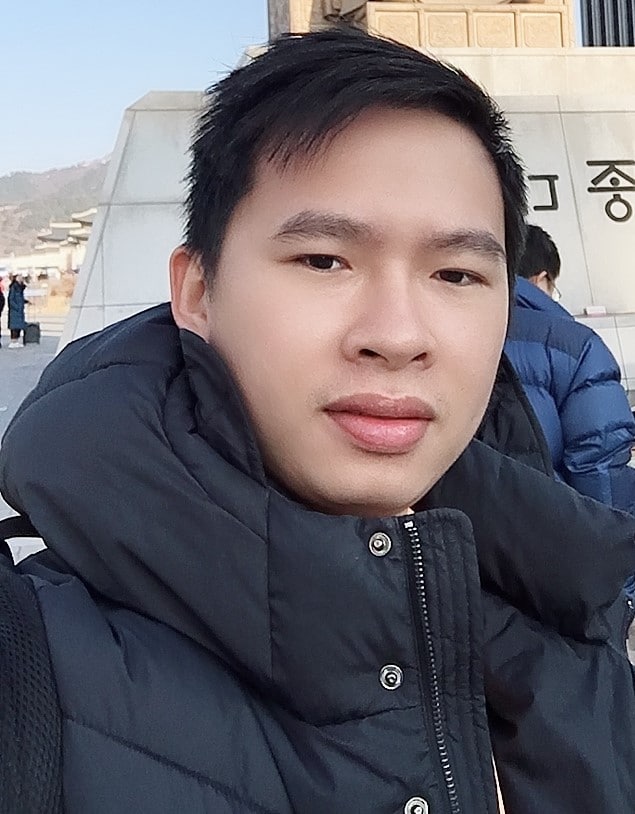}}]{\textbf{Thanh Nguyen}} (Graduate Student Member, IEEE) He received B.Sc degree in electronic and automation engineering from Ho Chi Minh City University of Science and Technology, in 2015. He has been pursuing an M.Sc. and Ph.D. degree at the Korea Advanced Institute of Science and Technology since 2018. His research interests include machine learning, deep learning, and reinforcement learning.
\end{IEEEbiography}
\begin{IEEEbiography}[{\includegraphics[width=1in,height=1.25in,clip,keepaspectratio]{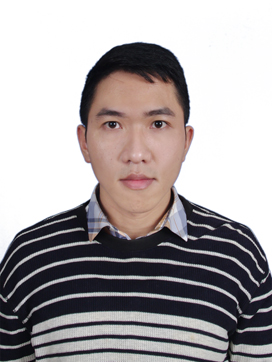}}]{\textbf{Tung M. Luu}} (Graduate Student Member, IEEE) He received B.Sc degree in electronic and telecommunication engineering from Hanoi University of Science and Technology, in 2017, and the M.Sc. degree in electrical engineering from Korea Advanced Institute of Science and Technology, in 2020. He is currently pursuing a Ph.D. degree with Korea Advanced Institute of Science and Technology. His research interests include machine learning, deep learning, and reinforcement learning.
\end{IEEEbiography}
\begin{IEEEbiography}[{\includegraphics[width=1in,height=1.25in,clip,keepaspectratio]{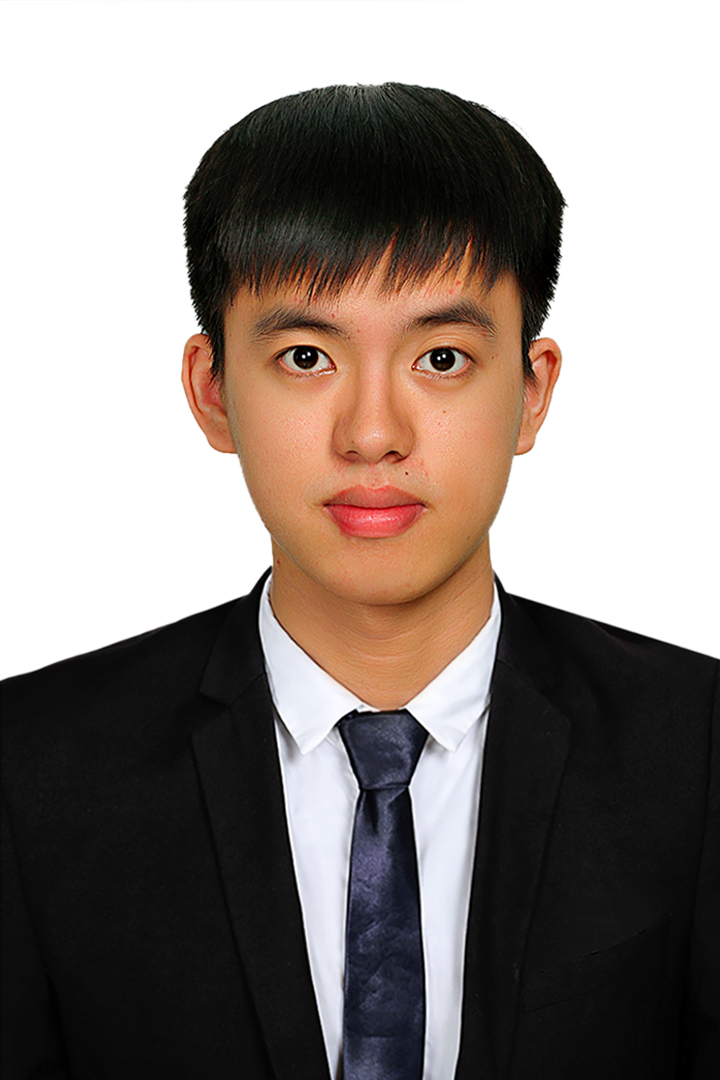}}]{\textbf{TRI TON}} (Graduate Student Member, IEEE) He received a B.Sc degree in Automation and Control Engineering from Ho Chi Minh University of Technology in 2021. He has been pursuing an M.Sc and Ph.D. degree at the Korea Advanced Institute of Science and Technology since 2022. His research interests are in computer vision, specifically focusing on 3D understanding.
\end{IEEEbiography}
\begin{IEEEbiography}
[{\includegraphics[width=1in,height=1.25in,clip,keepaspectratio]{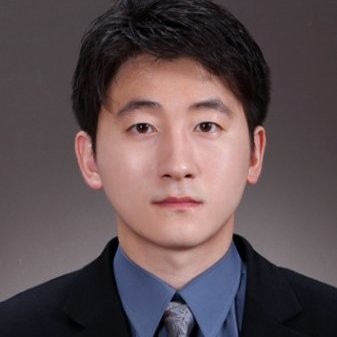}}]{\textbf{SungWoong Kim}} (Senior Member, IEEE) He is an Associate Professor in the department of Artificial Intelligence at Korea University. He directs AGI Lab where the research focuses on realizing artificial general intelligence to make an AI agent can perform task generalization and self-learning. He received BS and Ph.D degrees from KAIST in 2004 and 2011, respectively. When he was a graduate student, his research area was machine learning, especially applied to speech and image processing, under supervision by Professor Chang D. Yoo. In the middle of his Ph.D course, he performed research internships at National ICT Australia under supervision by Dr. Alex Smola and Mircosoft Research Cambridge under supervision by Dr. Pushmeet Kohli and Dr. Sebastian Nowozin. After receiving his Ph.D degree, he was in KAIST as a postdoc researcher, and then worked as a staff engineer at Qualcomm Research Korea. In 2017, He joined Kakao Brain and worked as a research scientist conducting several research projects mostly related to artificial general intelligence. Then, in March 2023, He joined Korea University.
\end{IEEEbiography}
\begin{IEEEbiography}[{\includegraphics[width=1in,height=1.25in,clip,keepaspectratio]{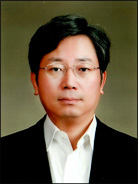}}]{\textbf{Chang D. Yoo}} (Senior Member, IEEE) He received the B.S. degree in Engineering and Applied Science from the California Institute of Technology, the M.S. degree in Electrical Engineering from Cornell University and the Ph.D. degree in Electrical Engineering from the Massachusetts Institute of Technology. From January 1997 to March 1999 he was Senior Researcher at Korea Telecom (KT). Since 1999, he has been on the faculty at the Korea Advanced Institute of Science and Technology (KAIST), where he is currently a Full Professor with tenure in the School of Electrical Engineering and an Adjunct Professor in the Department of Computer Science. He also served as Dean of the Office of Special Projects and Dean of the Office of International Relations, respectively.
\end{IEEEbiography}
\EOD

\end{document}